\documentclass[10pt,twocolumn,letterpaper]{article}

\usepackage{cvpr}              % To produce the CAMERA-READY version
\usepackage{graphicx}
\usepackage{amsmath}
\usepackage{amssymb}
\usepackage{booktabs}

\usepackage[table]{xcolor}
\usepackage[british, english, american]{babel}
\usepackage{makecell}
\usepackage{multirow}
\usepackage{pifont}
\usepackage[accsupp]{axessibility}  % Improves PDF readability for those with disabilities.

\newcommand{\tablestyle}[2]{\setlength{\tabcolsep}{#1}\renewcommand{\arraystretch}{#2}\centering\footnotesize}
\newcommand{\cmark}{\ding{51}}%
\newcommand{\xmark}{\ding{55}}%
\renewcommand{\paragraph}[1]{\vspace{1.25mm}\noindent\textbf{#1}}

\newlength\savewidth\newcommand\shline{\noalign{\global\savewidth\arrayrulewidth
  \global\arrayrulewidth 1pt}\hline\noalign{\global\arrayrulewidth\savewidth}}
\usepackage[pagebackref=false, breaklinks=true, letterpaper=true, colorlinks,
            citecolor=citecolor, linkcolor=linkcolor, bookmarks=false]{hyperref}
\definecolor{citecolor}{HTML}{0071BC}
\definecolor{linkcolor}{HTML}{ED1C24}
\definecolor{deemph}{gray}{0.6}
\definecolor{mygray}{gray}{.9}
\definecolor{mypink}{rgb}{.99,.91,.95}
\definecolor{mycyan}{cmyk}{.1,0,0,0}
\definecolor{capri}{rgb}{0.0, 0.75, 1.0}
\definecolor{brightcerulean}{rgb}{0.11, 0.67, 0.84}
\newcommand{\gc}[1]{\textcolor{deemph}{#1}}

\usepackage{amsmath,amsfonts,bm}

\def\eqref#1{equation~\ref{#1}}
\def\1{\bm{1}}

\def\vx{{\bm{x}}}

\DeclareMathAlphabet{\mathsfit}{\encodingdefault}{\sfdefault}{m}{sl}
\SetMathAlphabet{\mathsfit}{bold}{\encodingdefault}{\sfdefault}{bx}{n}

\def \xi{\vx^I}

\DeclareMathOperator*{\argmax}{arg\,max}

\usepackage[capitalize]{cleveref}
\crefname{section}{Sec.}{Secs.}
\Crefname{section}{Section}{Sections}
\Crefname{table}{Table}{Tables}
\crefname{table}{Tab.}{Tabs.}

\def\cvprPaperID{4914} % *** Enter the CVPR Paper ID here
\def\confName{CVPR}
\def\confYear{2023}

\definecolor{baselinecolor}{gray}{.9}
\newcommand{\baseline}[1]{\cellcolor{baselinecolor}{#1}}

\begin{document}

%%%%%%%%% TITLE - PLEASE UPDATE
\title{
\vspace{-8mm}
DetCLIPv2: Scalable Open-Vocabulary Object Detection Pre-training via Word-Region Alignment
\vspace{-5mm}}

%DetCLIPv2: Enabling Large-scale Image-text Pair Pre-training for Open-world Detection with Fine-grained Proposal-to-class Alignment

\author{%
  Lewei Yao$^{1,2}$, Jianhua Han$^2$, Xiaodan Liang$^3$\footnotemark[2] , Dan Xu$^1$, \\Wei Zhang$^2$, Zhenguo Li$^2$, Hang Xu$^2$\footnotemark[2]  \\
  \small{$^1$Hong Kong University of Science and Technology, $^2$Huawei Noah's Ark Lab} \\
  \small {$^3$Shenzhen Campus of Sun Yat-Sen University }
}

\twocolumn[{%
\renewcommand\twocolumn[1][]{#1}%
\maketitle
\vspace{-12mm}
\begin{center}
    \centering
    \includegraphics[width=\textwidth]{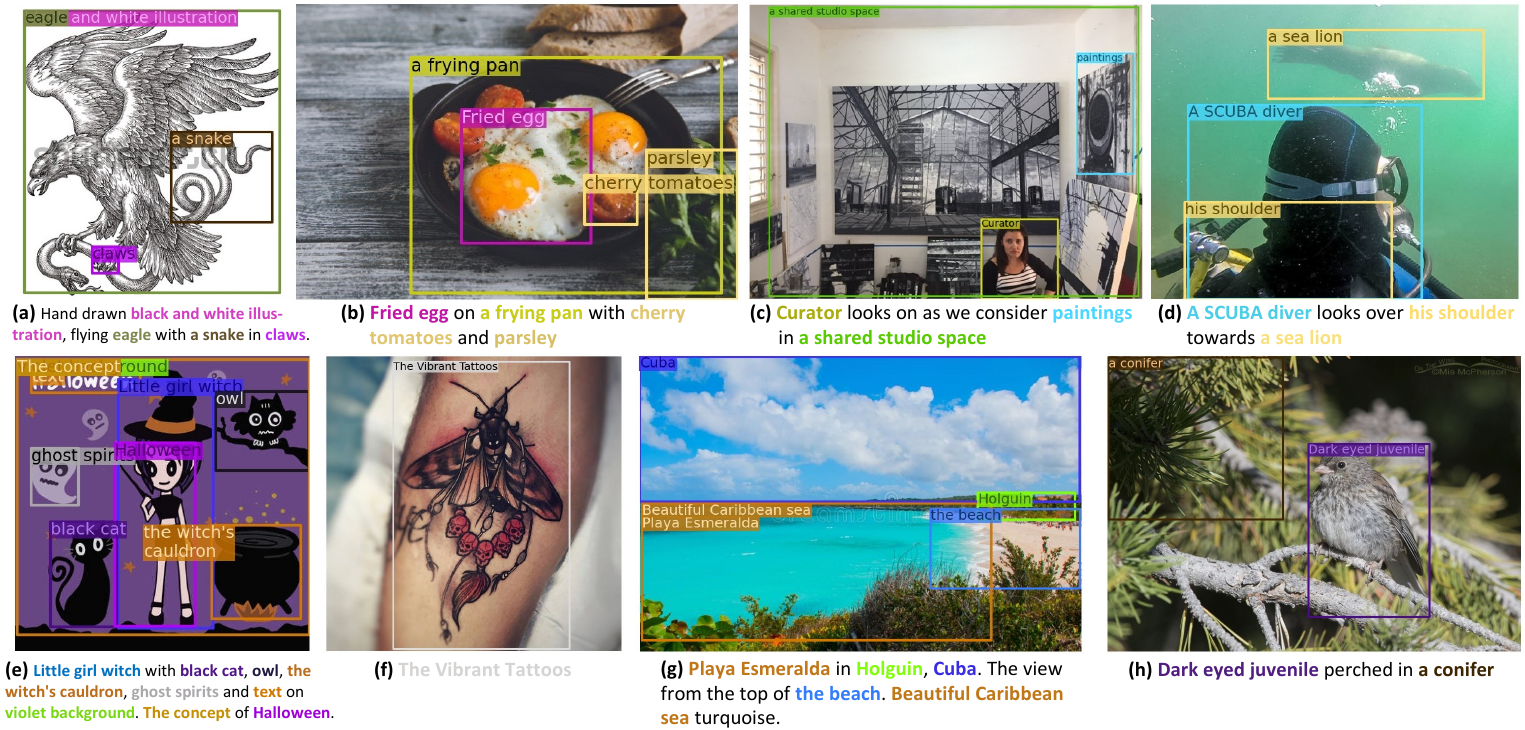}  
\vspace{-7mm}
\captionof{figure}{\textbf{Visualizations of DetCLIPv2 for open-vocabulary word-region alignment}. DetCLIPv2 is able to detect broad concepts.
\vspace{-1mm}
% with wide domains.
}
\label{fig:word_region_align}
\maketitle
\end{center}%
}]

\maketitle

\renewcommand{\thefootnote}{\fnsymbol{footnote}}
\footnotetext[2]{\vspace{-0.5mm}Corresponding author: liangxd9@mail.sysu.edu.cn, xu.hang@huawei.com}
\renewcommand{\thefootnote}{\arabic{footnote}}

\begin{abstract}
\vspace{-0.5em}
This paper presents DetCLIPv2, an efficient and scalable training framework that incorporates large-scale image-text pairs to achieve open-vocabulary object detection (OVD). Unlike previous OVD frameworks that typically rely on a pre-trained vision-language model (e.g., CLIP) or exploit image-text pairs via a pseudo labeling process, DetCLIPv2 directly learns the fine-grained word-region alignment from massive image-text pairs in an end-to-end manner. To accomplish this, we employ a maximum word-region similarity between region proposals and textual words to guide the contrastive objective. To enable the model to gain localization capability while learning broad concepts, DetCLIPv2 is trained with a hybrid supervision from detection, grounding and image-text pair data under a unified data formulation. By jointly training with an alternating scheme and adopting low-resolution input for image-text pairs, DetCLIPv2 exploits image-text pair data efficiently and effectively: DetCLIPv2 utilizes 13$\times$ more image-text pairs than DetCLIP with a similar training time and improves performance. With 13M image-text pairs for pre-training, DetCLIPv2 demonstrates superior open-vocabulary detection performance, e.g., DetCLIPv2 with Swin-T backbone achieves 40.4\% zero-shot AP on the LVIS benchmark, which outperforms previous works GLIP/GLIPv2/DetCLIP by 14.4/11.4/4.5\% AP, respectively, and even beats its fully-supervised counterpart by a large margin.
\end{abstract}
\vspace{-4mm}
\section{Introduction}
\label{sec:intro}

Traditional object detection frameworks~\cite{redmon2016you,ren2015faster,carion2020end,zhu2020deformable} are typically trained to predict a set of predefined categories, which fails to meet the demand of many downstream application scenarios that require to detect arbitrary categories (denoted as open-vocabulary detection, OVD). For example, a robust autonomous driving system requires accurate predictions for all classes of objects on the road~\cite{li2022coda}. Extending traditional object detectors to adapt these scenarios needs tremendous human effort for extra instance-level bounding-box annotations, especially for rare classes. To obtain an open-vocabulary detector without the expensive annotation process, the central question we should ask is:  \textit{where does knowledge about unseen categories come from}?

Recent works \cite{zang2022open,xie2021zsd,gu2021open} try to achieve open-vocabulary object detection by transferring knowledge from a pre-trained vision-language (VL) model \cite{radford2021learning, jia2021_align,yao2021filip}. E.g., ViLD~\cite{gu2021open} distills the CLIP's \cite{radford2021learning} image embeddings of cropped proposals into the proposal features of a detection model. 
However, these solutions suffer from the domain gap problem: VL models are typically pre-trained with an image-level supervision using a fixed resolution input, which are not capable of recognizing objects with various scales in the  detection task, especially for small objects.

Another line of work resorts to exploiting massive image-text pairs crawled from the Internet. To utilize the image-text pair data without instance-level annotation, approaches \cite{gao2021towards,li2021grounded, fontanel2022detecting,inkawhich2022self,yao2022detclip,regionclip} generate pseudo-bounding-box labels following a self-training paradigm \cite{sohn2020simple} or based on a pre-trained VL model \cite{radford2021learning}. However, their final performance is restricted by the quality of pseudo-labels provided by a detector trained with limited human-annotated concepts or a VL model suffering from the aforementioned domain gap problem. Besides, using high-resolution inputs similar to detection data for massive image-text pairs will impose a huge computational burden on training, preventing us from further scaling up image-text pairs. 

To address the above issues, we present DetCLIPv2, an end-to-end open-vocabulary detection pre-training framework that effectively incorporates large-scale image-text pairs. DetCLILPv2 simultaneously learns localization capability and knowledge of broad concepts without relying on a teacher model to provide pseudo labels. Specifically, we perform joint training with heterogeneous data from multiple sources, including detection \cite{shao2019objects365}, grounding \cite{kamath2021mdetr} and image-text pairs \cite{sharma2018conceptual, changpinyo2021cc12m}, under a unified data formulation. 
To enable image-text pairs without instance-level annotations to facilitate learning of detection, inspired by \cite{yao2021filip}, we employ an optimal matching-based set similarity between visual regions and textual concepts to guide the contrastive learning. By alternating different types of data for training, we enable a “flywheel effect”: learning from detection data provides accurate localization, which helps extract representative regions for contrastive learning, while contrastive learning from image-text pairs helps recognize broader concepts, which further improves the localization of unseen categories. As the training goes on, the detector learns to locate and recognize increasingly rich concepts.

Furthermore, to relief the computation burden brought by large-scale image-text pairs, we adopt a low-resolution input for image-text pair data, which significantly improves the training efficiency. This is a reasonable design since the caption of image-text pair data typically describes only the main objects appearing in the image, which alleviates the necessity of high-resolution training.

Benefiting from the effective designs, DetCLIPv2 demonstrates superior open-vocabulary detection performance and promising scaling behavior. E.g., compared to the prior work DetCLIP \cite{yao2022detclip}, DetCLIPv2 is able to exploit 13$\times$ more image-text pairs while requiring only a similar training time. Using the vanilla ATSS \cite{zhang2020bridging} as the detector, DetCLIPv2 with Swin-T backbone achieves 40.4\% \textit{zero-shot} AP on the LVIS \cite{gupta2019lvis} benchmark, surpassing previous works GLIP\cite{li2021grounded}/GLIPv2 \cite{zhang2022glipv2}/DetCLIP \cite{yao2022detclip} by 14.4/11.4/4.5\% AP, respectively. DetCLIPv2 also exhibits great generalization when transferring to down-stream tasks, e.g., it achieves SoTA fine-tuning performance on LVIS and ODinW13 \cite{li2021grounded}. We present a possibility of achieving open-world detection by incorporating large-scale image-text pairs and hope it will enlighten the community to explore a similar successful trajectory to CLIP \cite{radford2021learning}.

\begin{figure}
\begin{center}
\includegraphics[width=\linewidth]{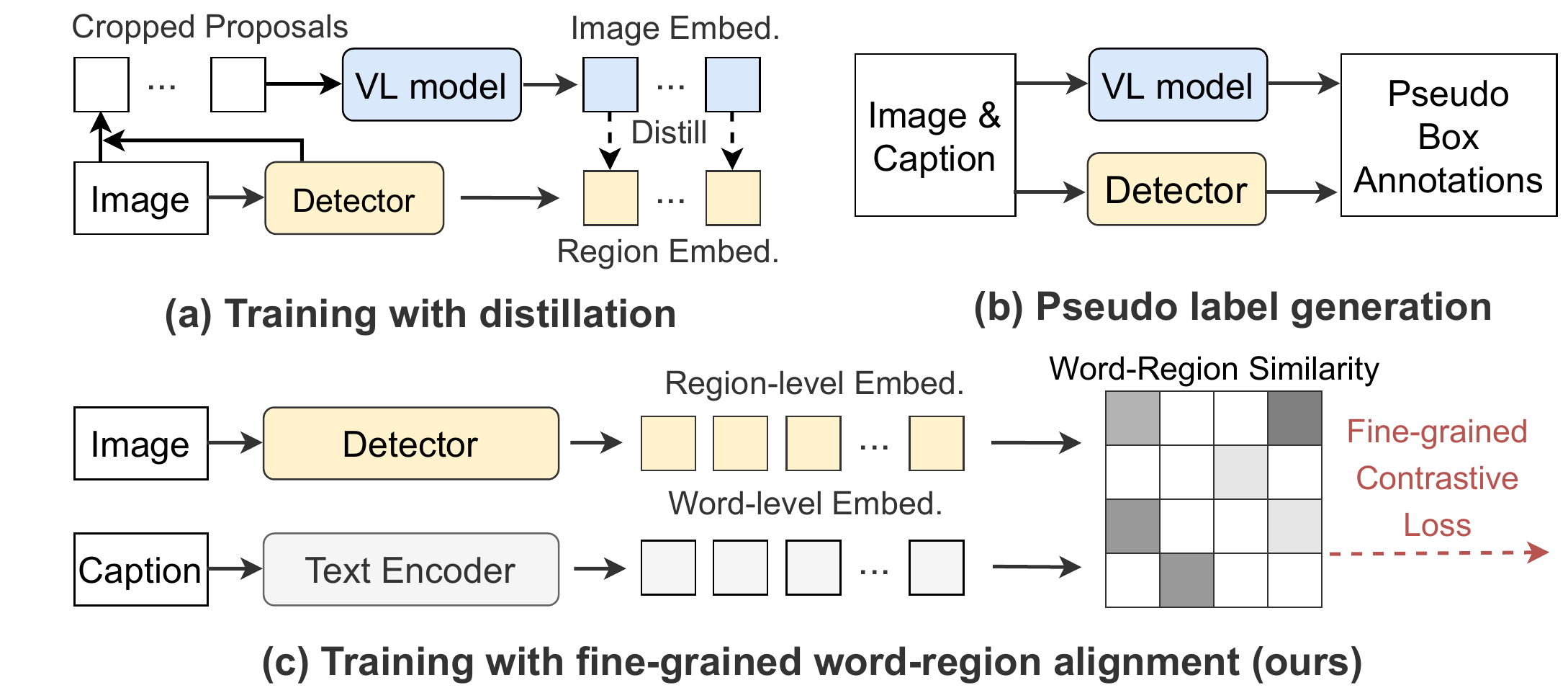}
\vspace{-6mm}
\caption{\textbf{Different OVD training paradigms.} (a) Distilling knowledge from a pre-trained VL model \cite{gu2021open}. (b) Exploiting image-text pairs via pseudo labeling \cite{li2021grounded}. (c) Our end-to-end joint training eliminates complex multi-stage training schemes, allowing for mutual benefits in learning from different types of data.
}
\label{fig:introduction}
\end{center}
\vspace{-7mm}
\end{figure}

\vspace{-1mm}
\section{Related Work}
\vspace{-1mm}
\label{sec:related_work}

\paragraph{Vision-Language Pre-training (VLP).} Conventional vision-language models are designed to serve a specific task, e.g., VQA \cite{kim2021vilt,antol2015vqa,li2020oscar,gao2019dynamic} and image captioning \cite{xu2015show,anderson2018bottom,yao2018exploring,liu2021cptr}, etc. Recently, there has been a trend to develop generic vision-language representation learning systems by exploiting large-scale low-cost image-text pairs. For example, CLIP \cite{radford2021learning} and ALIGN \cite{jia2021_align} perform cross-modal contrastive learning on millions of image-text pairs and achieve impressive zero-shot image classification performance. The most relevant work to our approach is FILIP \cite{yao2021filip}, which proposes a cross-modal late interaction  mechanism based on a word-patch similarity to better facilitate image-text alignment. However, it is non-trivial to leverage the idea to construct an open-vocabulary detection system, for which our approach provides a solution.

\paragraph{Open-vocabulary object detection (OVD)} emerges recently as a
more general and practical paradigm to detect objects of unbounded concepts. Inspired by the success of vision-language pre-training, recent works \cite{zang2022open,xie2021zsd,gu2021open,regionclip} propose to transfer knowledge of a pre-trained VL model (e.g., CLIP \cite{radford2021learning}) into a detector. 
Another effective idea is to use a wider source of training data. E.g, \cite{gao2021towards,li2021grounded, fontanel2022detecting,inkawhich2022self,yao2022detclip} incorporate low-cost image-text pairs to expand domain coverage via a pseudo labeling process. XDETR \cite{cai2022x} integrates a standard contrastive learning in VLP \cite{radford2021learning,jia2021_align} to learn image-to-text alignment. Detic \cite{zhou2022detecting} turns to solve a large-vocabulary detection problem by  directly assigning classification labels to the max-size region proposals. Unlike previous works, our approach targets on building an end-to-end framework that effectively learns word-region alignment from massive image-text pairs without relying on a teacher model.

\paragraph{Semi-supervised Object Detection (SSOD)} methods \cite{sohn2020simple,zoph2020rethinking,tang2021proposal,xu2021end} aim to improve object detection systems by exploiting unlabeled data on the basis of some available labeled data. Although effective in improving performance, these methods still assume a closed-domain setting where the categories in unlabeled data should be covered by labeled data. On the other hand, \textbf{Weakly-supervised Object Detection (WSOD)} methods \cite{bilen2016weakly,chen2020slv,mai2020erasing} aim to establish localization-capable detectors by leveraging image-level labels, which also require a set of pre-defined categories. Differing from methods in these fields, our approach considers a more challenging open-domain setting and targeting on establishing an open-world detector by learning unlimited concepts from massive image-text pairs.

\vspace{-1mm}
\section{The Proposed Approach}
\vspace{-1mm}
%############################################################################################
% framework figure
\begin{figure*}
\vspace{-4mm}
\begin{center}
\includegraphics[width=\textwidth]{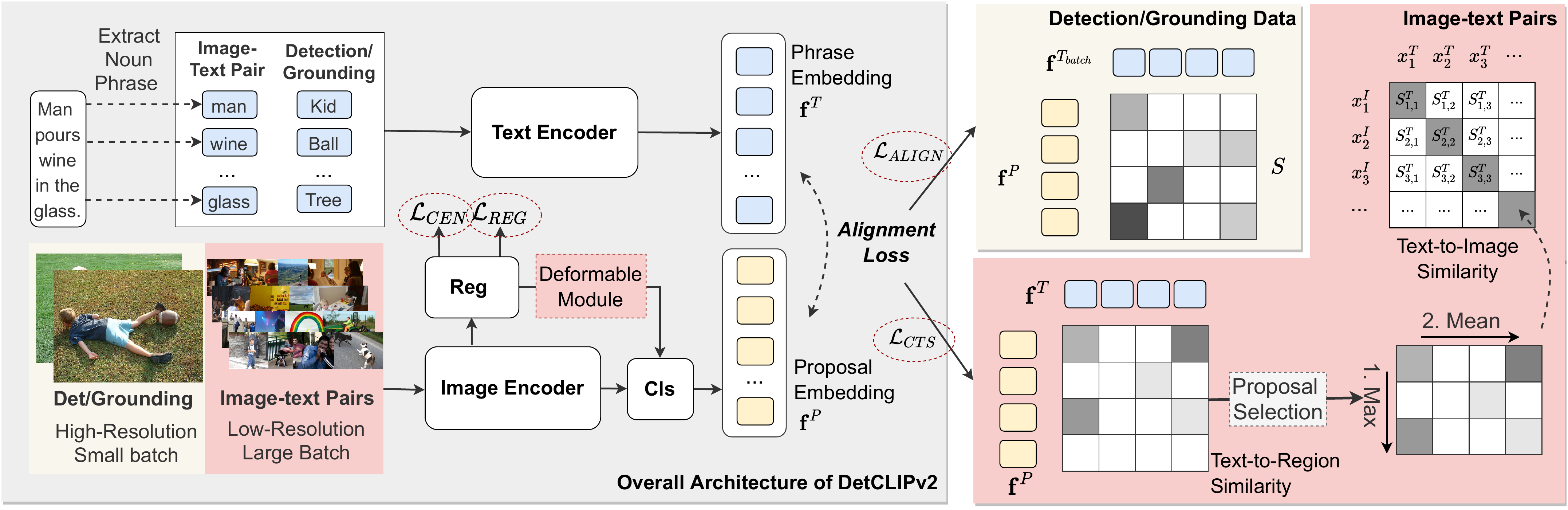}
\vspace{-6mm}
\caption{\textbf{Overall architecture of DetCLIPv2}. DetCLIPv2 performs a joint training with detection, grounding and image-pair data in an end-to-end manner. 
The architecture consists of an image encoder to extract region embeddings $\textbf{f}^P$ from an input image and a text encoder to compute word embeddings $\textbf{f}^T$ for the input noun phrases. For detection and grounding data, the learning is performed by aligning the word-region similarity matrix $S$ to a target matrix constructed with instance-level annotations. For image-text pairs, we calculate an optimal match-based set similarity between $\textbf{f}^T$ and $\textbf{f}^P$ to guide the contrastive learning, enabling the learning of word-region alignment.
}
% patches in the images and words in the sentences.}
\label{fig:framework}
\end{center}
\vspace{-7mm}
\end{figure*}
%############################################################################################

%############################################################################################
% a paragraph of overview: goal/ input,output

An overview framework of the proposed approach is illustrated in Figure~\ref{fig:framework}. To construct a robust open-world object detection system, DetCLIPv2 incorporates data from different sources, i.e., detection, grounding, and image-text pairs, for pre-training. We first introduce a unified paralleled data formulation enabling a training with heterogeneous supervisions (Sec.~\ref{data_fomulation}). To utilize image-text pairs without instance-level annotations, we introduce a fine-grained contrastive learning that automatically aligns textual words and visual regions (Sec.~\ref{contrastive_learning}). Finally, we introduce the model architecture/training objective (Sec.~\ref{objective_and_model}) and the joint-training details (Sec.~\ref{joint_training}).

%############################################################################################
\vspace{-1mm}
\subsection{A Unified Data Formulation} \label{data_fomulation}
\vspace{-1mm}
Following DetCLIP \cite{yao2022detclip}, we use a \textit{paralleled formulation} to unify the formats of data from different sources. Specifically, we formulate each data sample as a triplet: $(x^I, \{\textbf{b}_i\}_{i=1}^N, \{t_j\}_{j=1}^M )$, where $x^I \in \mathbb{R}^{3\times h \times w}$ is the image, $\{\textbf{b}_i| \textbf{b}_i\in \mathbb{R}^4\}_{i=1}^N$ and $T=\{t_j\}_{j=1}^M$ denote a set of bounding box annotations and concept names, respectively. The triplet is constructed for different types of data differently:
\begin{itemize}
    \item \textbf{Detection}. $T$ is constructed from a sampled category names of the dataset, which consists of categories appearing in the image and additional randomly-sampled negative categories.
    To explicitly provide the relationships between various concepts, We apply \textit{concept enrichment} \cite{yao2022detclip} during both training and testing phases, i.e., each $t_j$ is obtained by concatenating its category name with the corresponding definition.
    \vspace{-2mm}
    \item \textbf{Grounding}. We first extract noun phrases (provided in annotations) from the original caption to form a positive concept set $T_{pos}=\{t_j\}_{j=1}^{|pos|}$. To provide enough negative concepts for learning, we further randomly sample a negative concept set  $T_{neg}=\{t_j\}_{j=1}^{|neg|}$ that does not contained in the caption (i.e., $T_{pos}\cap T_{neg}=\emptyset$) from a constructed
    \textit{concept dictionary} \cite{yao2022detclip}. The final category name set is formed by $T=T_{pos} \cup T_{neg}$. 
    \vspace{-2mm}
    \item \textbf{Image-text pairs}. As instance-level annotation is not available, we have $\{b_i\}_{i=1}^N=\emptyset$. $T$ consists of the original caption and noun phrases extracted from it\footnotemark.
\end{itemize}

\footnotetext{We use NLP parser provided by Spacy \cite{spacy} repository.}

For detection and grounding data, each $\textbf{b}_i$ is labeled with a concept $t_j$, which enables the learning of open-vocabulary object detection. We describe it as follows.

\paragraph{Open-vocabulary object detection.} As illustrated in Figure~\ref{fig:framework}, we use a dual-stream architecture which consists of an image encoder and a text encoder. The image encoder is an arbitrary-form object detector that takes an image $x^I$ as the input and outputs a set of region proposals $P=\{\textbf{p}_k\}_{k=1}^{K}$ (for one-stage detector, $K$ equals to number of anchors), as well as their classification features $\textbf{f}^P\in \mathbb{R}^{K\times D}$, where $D$ is the feature dimension. For the text side, we treat each concept name $t_j$ as a sentence and forward all $t_j \in T$\ to the text encoder \textit{separately} to obtain the sentence embeddings $\textbf{f}^T\in \mathbb{R}^{M\times D}$. Following previous works \cite{radford2021learning,jia2021_align,zhang2022glipv2}, to increase the number of negative samples, we collect $\textbf{f}^T$ across a global batch and remove duplicate concepts contained in different samples in a batch, which gives a gathered text embedding $\textbf{f}^{T_{batch}}\in\mathbb{R}^{M_B\times D}$, where $M_B$ is the total number of concepts in a global batch after deduplication. Then we calculate the similarity matrix $S\in \mathbb{R}^{K\times M_B}$ between $\textbf{f}^P$ and $\textbf{f}^{T_{batch}}$ by

\vspace{-.6em}
\begin{equation}
\label{eq:similarity}
S = \textbf{f}^P (\textbf{f}^{T_{batch}})^\top
\vspace{-.2em}
\end{equation}

When instance-level annotations are available, e.g., for detection and grounding data, we can construct a target matrix $G\in \{0,1\}^{K\times M_B}$ following a ground-truth assignment process in conventional object detection frameworks \cite{ren2015faster,zhang2020bridging,Tian_fcos}, then the alignment loss $\mathcal{L}_{align}(S,G)$ (detailed in Sec. \ref{objective_and_model}) can be calculated;  while for image-text pairs where the instance-level annotation is not available, we elaborate our approach in the Sec.~\ref{contrastive_learning}. 
\vspace{-1mm}
\subsection{Learning from Image-text Pairs} \label{contrastive_learning}
\vspace{-1mm}
Massive image-text pairs crawled from the Internet can provide rich knowledge for the learning of visual-language models. However, due to the lack of instance-level annotations, it is non-trivial to leverage image-text pairs to improve a dense prediction (e.g, object detection) learning system. Inspired by \cite{yao2021filip}, we introduce a contrastive learning method to learn a fine-grained word-region correspondences without relying on instance-level annotation, which is described as follows.

\paragraph{Word-region alignment similarity.}
Given an image-text pair $(x^I,x^T)$, we extract a set of noun phrases $T=\{t_j\}_{j=1}^M$ from $x^T$ and take $(x^I,\{t_j\}_{j=1}^M)$ as the input of the model. The image encoder generates a set of proposals $P=\{\textbf{p}_k\}_{k=1}^{K}$ from $x^I$ with their region features $\textbf{f}^P\in \mathbb{R}^{K\times D}$ and the text encoder extracts text embeddings $\textbf{f}^T\in \mathbb{R}^{M\times D}$ of $\{t_j\}_{j=1}^M$. Our word-region alignment contrastive learning is constructed based on the set similarity between $P$ and $T$. Specifically, for $j$-th concept $t_j\in T$, we find its closest match in $P$ by calculating
\vspace{-.2em}
\begin{equation}
\label{eq:token-wise similarity}
m_j=\argmax_{0< k \le K} [\textbf{f}^T]_j^\top [\textbf{f}^P]_k,
\vspace{-.2em}
\end{equation}
where $[\textbf{f}^P]_k$ is the $k$-th region feature in $\textbf{f}^P$, and similar for $[\textbf{f}^T]_j$. 
This operation can be interpreted as, for each concept we find a region that best fits its description. Then we calculate the text-to-image similarity $s^T$ between $x^I$ and $x^T$ by aggregating all word-to-region similarities, i.e.,
\vspace{-.2em}
\begin{equation}
\label{eq:token-wise similarity}
s^T(x^I,x^T)= \frac{1}{M}\sum_{j=1}^{M} [\textbf{f}^T]_j^\top [\textbf{f}^P]_{m_j}
\vspace{-.4em}
\end{equation}

Note that the image-to-text similarity $s^I(x^I,x^T)$ can be calculated in a similar way. However, we exclude this part from our algorithm, since image-text pairs crawled from the Internet suffer from a severe \textit{partial labeling} problem -- for the vast majority of data, the text describes only a small fraction of the objects appearing in the image, i.e., most of region proposals cannot find their corresponding match in the caption texts. Including image-to-text matching can result in a noticeable performance degradation, for which we give an ablation in Sec.~\ref{subsubsec:contrastive_ablation}.

Another reasonable consideration is that each textual concept should correspond to multiple regions. This design can be modeled by using a softmax-weighted-sum similarity between a textual concept and all visual regions, i.e.,
\vspace{-.2em}
\begin{equation}
\label{eq:weighted_sum similarity}
s^T(x^I,x^T) = \frac{1}{M} \sum_{j=1}^{M} \sum_{k=1}^{P} \frac{\exp(s_{j,k}/\tau_t)}{\sum_{i=1}^{P}\exp(s_{j,i}/\tau_t)}s_{j,k}
\vspace{-.4em}
\end{equation}
where $s_{j,k}=[\textbf{f}^T]_j^\top [\textbf{f}^P]_k$ is the similarity between $j$-th textual concept and $k$-th visual region, and $\tau_t$ is a temperature hyper-parameter to control sharpness of the softmax-based weights (when $\tau_t\rightarrow0$, Eq.~\ref{eq:weighted_sum similarity} degrades to Eq.~\ref{eq:token-wise similarity}). We investigate this design in Sec.~\ref{subsubsec:contrastive_ablation}.

\paragraph{Image-text contrastive loss.} Based on the introduced word-region alignment similarity, a standard contrastive learning between image-text pairs can be performed \cite{radford2021learning}. Specifically, assume a batch of $B$ image-text pairs $\{(x_i^I,x_i^T)\}_{i=1}^B$,
the contrastive loss $\mathcal{L}_{cts}$ is formulated as 
\vspace{-.2em}
\begin{equation}
\label{eq:contrastive_loss}
\mathcal{L}_{cts}=\mathcal{L}_{T\rightarrow I}=-\frac{1}{B} \log \frac{\exp(s^T(x_i^I,x_i^T)/\tau)}{\sum_{j=1}^{B} \exp(s^T(x_j^I,x_i^T)/\tau)}
\vspace{-.2em}
\end{equation}
where $s^T(x_i^I,x_j^T)$ is text-to-image word-region alignment similarity between $i$-th image $x_i^I$ and $j$-th text $x_j^T$, which is given by Eq.~\ref{eq:token-wise similarity}, and $\tau$ is a temperature to scale the logits. As discussed before, we only consider text-to-image contrastive loss. By incorporating the word-region alignment similarity, the contrastive loss helps the model learn fine-grained word-region correspondences automatically.

\paragraph{Proposal selection.} Intuitively, we expect to select the most representative regions in an image to calculate similarities with textual concepts. There are several schemes to accomplish this. For example, many detectors incorporate class-agnostic object scores in their designs, e.g., foreground classification score in RPN\cite{ren2015faster}, centerness in FCOS\cite{Tian_fcos}, etc., which 
can be utilized to generate high-quality region proposals with good generalization \cite{gu2021open,kim2022oln}. However, these approaches fail to take the textual information into consideration. To select regions valuable for contrastive learning, for each candidate region, we calculate its similarities with all textual concepts within a local batch, and use the maximum similarity as its objectness score. The benefits of this design are two-fold: (1) it selects the regions most relevant to the text description; (2) it selects hard negative concepts that described in other texts which may benefit the contrastive learning. With the objectness score, we select top-$k$ 
proposals after a NMS operation. Different proposal selection strategies and the optimal $k$ are studied in Sec.~\ref{subsubsec:contrastive_ablation}.

\subsection{Model Architecture and Training Objective} \label{objective_and_model}
\vspace{-1mm}
\paragraph{Model architecture.} Similar to DetCLIP\cite{yao2022detclip}, DetCLIPv2 is built using the vanilla ATSS \cite{zhang2020bridging} detector equipped with a transformer-based \cite{radford2019language,radford2021learning} text encoder. 
We do not introduce additional heavy modules such as DyHead \cite{dai2021dynamic} adopted in \cite{li2021grounded,zhang2022glipv2} and cross-modal fusion adopted in \cite{li2021grounded,zhang2022glipv2,dou2022coarse}. 

A special design is that we insert a lightweight deformable convolution \cite{zhu2019deformable} at the beginning of the classification head, which uses the features output by the regression head to calculate the spatial offsets and the modulation scalar, and aggregates the features from the backbone output.  The motivation is that when training with image-text pairs, there is no supervision signal on the regression branch and therefore no gradient is generated. This design helps the gradient from the classification head to flow back to the regression head, so that the regression head also benefits from training with massive image-text pairs. I.e., learning a better spatial aggregation for backbone features helps regression head acquire better localization ability. We show this neat design provides substantial performance improvement when training with image-text pairs (see Sec.~\ref{subsubsec:deformable}).

\paragraph{Training Objective.} The overall objective of DetCLIPv2 can be formulated as 
\vspace{-.2em}
\begin{equation}
\label{eq:overall_objectives}
\mathcal{L}=\begin{cases}
\mathcal{L}_{align}+\alpha\mathcal{L}_{reg}+\beta\mathcal{L}_{center}, & \text{for detection} \\
\mathcal{L}_{align}, & \text{for grounding} \\
\lambda\mathcal{L}_{cts}, & \text{for image-text pairs} \\
\end{cases}
\vspace{-.2em}
\end{equation}
where $\mathcal{L}_{align}$ is the alignment loss described in Sec.~\ref{data_fomulation};  $\mathcal{L}_{cts}$ is the contrastive loss in Eq.~\ref{eq:contrastive_loss}; $\mathcal{L}_{reg}$ and $\mathcal{L}_{center}$ are regression and centerness losses, respectively; $\alpha$, $\beta$ and $\lambda$ are loss weights. Following ATSS \cite{zhang2020bridging}, we use focal loss for $\mathcal{L}_{align}$, GIoU loss \cite{rezatofighi2019generalized} for $\mathcal{L}_{reg}$, and cross-entropy loss for $\mathcal{L}_{center}$. We remove the localization loss for grounding data due to its inaccurate bounding box annotations.

\vspace{-2mm}
\subsection{Joint Training} \label{joint_training}
\vspace{-1mm}
% ###################################################################################
DetCLIPv2 performs a joint training with heterogeneous datasets. During training, we group data belonging to the same type for a global batch. At each iteration, we sample one type of data for training. Different data types are trained with different input resolutions and batch sizes. Specifically, we use a high-resolution input with a small batch size for detection and grounding data; while for image-text pairs, a low-resolution input with a large batch size is adopted, which helps increase the number of negative samples in contrastive learning and considerably reduce the training cost of massive image-text pairs.

\vspace{-2mm}
\section{Experimental Results}
\label{subsec:Experimental Results}
\vspace{-1mm}
\subsection{Implementation Details}\label{subsec:implementation_detail}
\vspace{-1mm}
\paragraph{Training Dataset.} We use multiple datasets from different sources for training (Table~\ref{tab:dataset_info}). Specifically, for detection data, we use a sampled subset from Objects365v2 \cite{shao2019objects365} dataset (denoted as O365) with 0.66M images; for grounding data, we use GoldG \cite{kamath2021mdetr} with COCO \cite{Lin2014coco} images removed, which results in a fairer zero-shot evaluation on LVIS \cite{gupta2019lvis}. For image-text pairs, we use 2 versions of Conceptual Captions (CC) datasets, i.e., CC3M \cite{sharma2018conceptual} and CC12M \cite{changpinyo2021cc12m} (together denoted as CC15M). 

% training dataset information
\begin{table}[t]
\centering
\resizebox{\linewidth}{!}{
\begin{tabular}{lcc}
\toprule
Dataset & Type & Volume \\
\midrule 
Objects365 \cite{shao2019objects365} (O365)  & Detection & 0.66M  \\
GoldG \cite{kamath2021mdetr} &Grounding & 0.77M  \\
\makecell[l]{CC15M \\(CC3M \cite{sharma2018conceptual}+CC12M \cite{changpinyo2021cc12m})} & Image-text pairs  & \makecell[c]{13M \\ (3M+10M)}  \\
\bottomrule
\end{tabular}
}
\vspace{-.7em}
\caption{A summary of training data. CC15M contains only 13M image-text pairs since some urls are invalid.}
\vspace{-1em}
\label{tab:dataset_info}
\end{table}

\paragraph{Training details.} We use Swin-transformer \cite{liu2021swin} backbones for image encoder. For text encoder, the maximum token length is set to 16 for efficient training and inference. We initialize the text-encoder with a pretrained FILIP model \cite{yao2021filip}. 32/64 V100 GPUs are used for training Swin-T/L-based models, respectively. For detection and grounding data, we use input resolution $1333\times800$ with a batch size of 128/256 for Swin-T/L models (4 per card), respectively; and for image-text pairs, we use input resolution $320\times320$ with a batch size of 6144 (192/96 per card for Swin-T/L model). We set $\alpha=2$ and  $\beta=1$ and $\lambda=0.1$ in Eq.~\ref{eq:overall_objectives}. Without otherwise specified, all models are trained with 12 epochs. More training details can refer to Appendix.

\paragraph{Evaluation benchmark.}
Following GLIP \cite{li2021grounded} and DetCLIP \cite{yao2022detclip}, we evaluate our method's \textit{zero-shot} performances on LVIS \cite{gupta2019lvis} with 1203 categories. \textit{Fixed} AP \cite{dave2021evaluating} on LVIS minival5k are reported for ablation and comparison with other methods. To further study the generalization ability of our method, we also evaluate with ODinW13 dataset \cite{li2021grounded,zhang2022glipv2}, which contains 13 downstream detection tasks with highly varied distributions. We focus on the GLIP protocol \cite{li2021grounded} rather than the ViLD protocol \cite{gu2021open} that splits LVIS into seen/unseen categories, since the former is a stronger and more practical open-world setting that does not make any prior assumptions on downstream tasks while the latter still requires partial LVIS data for training.

% ######################### Ablation studied #############################
\vspace{-1mm}
\subsection{Ablation Studies}\label{subsec:ablation_study}
\vspace{-1mm}
%##########################################################################
%overall table of contrastive learning ablations 
\begin{table*}[t]
\vspace{-4mm}
\centering
%#################################################
% proposal selection strategy
%#################################################
\subfloat[
\textbf{Proposal selection strategy}. Batch-wised text similarity generates better proposals for contrastive learning.
\label{tab:proposal_selection}
]{
\centering
\begin{minipage}{0.29\linewidth}{\begin{center}
\tablestyle{4pt}{1.05}
\begin{tabular}{lcc}
\# & Strategy & AP (r/c/f)\\
\shline
% \shline
1 & cls  & 28.4 (26.6/28.2/28.8)    \\
2 & IoU & 30.1 (30.0/30.1/30.2)   \\
3 & centerness & 30.2 (28.4/30.6/30.1)   \\
4 & text sim (S) &  29.6 (24.9/29.5/30.5) \\
5 & text sim (B) & \baseline{\textbf{31.3} (29.4/\textbf{31.7}/\textbf{31.3})} \\
6 & ~~+centerness & 30.9 (\textbf{30.2}/31.1/30.8) \\
\end{tabular}
\end{center}}\end{minipage}
 }
\hspace{2em}
%#################################################
% Word-region alignment strategy
%#################################################
\subfloat[
\textbf{Word-region matching strategy}. Matching each textual concept to the closest region is effective and memory-efficient.
\label{tab:matching_strategy}
]{
\centering
\begin{minipage}{0.29\linewidth}{\begin{center}
\tablestyle{1.5pt}{1.05}
\begin{tabular}{lccc}
\# &Strategy & AP (r/c/f) & Memory\\
\shline
1 & max-bbox & {29.8 (28.5/39.5/30.4)} &  19.8 GB \\
2 & 1-to-1 & \baseline{\textbf{31.3} (29.4/\textbf{31.7}/\textbf{31.3})} & 20.6 GB \\
3 & 1-to-many & {30.9 (\textbf{31.3}/30.7/31.1)} & 26.0 GB \\
\multicolumn{2}{c}{~}\\
\multicolumn{2}{c}{~}\\
\multicolumn{2}{c}{~}\\
\end{tabular}
\end{center}}\end{minipage}
}
\hspace{2em}
%#################################################
% contrastive loss design
%#################################################
\subfloat[
\textbf{Contrastive loss design}. Excluding image-to-text contrastive loss can boost the performance.
\label{tab:contrastive_loss}
]{
\begin{minipage}{0.29\linewidth}{\begin{center}
\tablestyle{4pt}{1.05}
\begin{tabular}{cc}
Design & AP (r/c/f)\\
\shline
text-to-image & \baseline{\textbf{31.3} (29.4/\textbf{31.7}/\textbf{31.3})}  \\
image-to-text & {29.8 (\textbf{30.0}/29.7/29.9)} \\
bilateral & {30.9 (\textbf{30.0}/31.5/30.5)} \\
\multicolumn{2}{c}{~}\\
\multicolumn{2}{c}{~}\\
\multicolumn{2}{c}{~}\\

\end{tabular}
\end{center}}\end{minipage}
}
\\
\centering
\vspace{.3em}

%#################################################
% num of proposals
%#################################################
\subfloat[
\textbf{Number of proposals}. We use $k=100$.
\label{tab:number_proposals}
]{
\begin{minipage}{0.29\linewidth}{\begin{center}
\tablestyle{6pt}{1.05}
\begin{tabular}{cc}

Top-k & AP (r/c/f) \\
\shline
25 & {30.6 (29.9/30.5/30.8)} \\
50 & {30.8 (\textbf{30.2}/30.6/31.0)} \\
100 & \baseline{\textbf{31.3} (29.4/\textbf{31.7}/\textbf{31.3})}  \\
200 & {30.8 (29.4/30.6/31.1)}  \\
\end{tabular}
\end{center}}\end{minipage}
}
\hspace{2em}
%#################################################
% temperature
%#################################################
\subfloat[
\textbf{Temperature $\tau$}. $\tau=0.5$ works the best.
\label{tab:temperature}
]{
\begin{minipage}{0.29\linewidth}{\begin{center}
\tablestyle{6pt}{1.05}
\begin{tabular}{cc}
$\tau$ & AP (r/c/f) \\
\shline
1 & 30.1 (28.4/30.3/30.3)  \\
0.5 & \baseline{\textbf{31.3} (29.4/\textbf{31.7}/\textbf{31.3})} \\
0.15 & 30.8 (\textbf{29.6}/31.0/30.9) \\
0.07 & 29.2 (27.5/29.0/29.6) \\
\end{tabular}
\end{center}}\end{minipage}
}
\hspace{2em}
%#################################################
% contrastive loss weight
%#################################################
\subfloat[
\textbf{Contrastive loss weight.} We use $\lambda=0.1$.
\label{tab:loss_weight}
]{
\begin{minipage}{0.29\linewidth}{\begin{center}
\tablestyle{6pt}{1.05}
\begin{tabular}{cc}
$\lambda$ & AP (r/c/f)\\
\shline
0.03 & {30.5 (\textbf{29.6/}30.0/31.0)}  \\
0.1 & \baseline{\textbf{31.3} (29.4/\textbf{31.7}/\textbf{31.3})} \\
0.3 & {30.9 (29.4/31.5/30.7)} \\
1 & {28.9 (27.7/28.8/29.3)} \\
\end{tabular}
\end{center}}\end{minipage}
}
%#################################################
\vspace{-1.5em}
\caption{\textbf{Ablation experiments for image-text contrastive learning}. The models are based on the Swin-T backbone and trained with O365+CC3M dataset. We report zero-shot \textit{fixed} AP (\%) \cite{dave2021evaluating} on LVIS minival5k \cite{kamath2021mdetr}. r/c/f indicate AP of rare/common/frequent categories, respectively. 
% Default settings are marked in \colorbox{baselinecolor}{gray}. 
Designs with higher overall AP (marked in \colorbox{baselinecolor}{gray}) are selected as our final setting. 
% Baseline of training with only O365 can be referred to the row1 of Table~\ref{tab:train_with_more_datasets}.
}
\label{tab:contrastive_ablations} \vspace{-1.5em}
\end{table*}
%###########################################################################################

%##############################################################################################
% effectiveness of deformable head
\begin{table}[t]
\centering
\tablestyle{1pt}{1.05}
\resizebox{\linewidth}{!}{
\begin{tabular}{lccc}
\toprule
Pretrain-data & deform & AP (r/c/f) & iter time (s) \\
\midrule
O365 & \xmark & {28.8 (26.0 / 28.0 / 30.0)} & 0.925 \\
O365 & \cmark & {28.6 (24.2 / 27.1 / 30.6)} & 1.075 \\
O365+GoldG+CC3M & \xmark & {37.3 (34.1 / 36.9 / 38.2)} &0.980\\
O365+GoldG+CC3M & \cmark & {\textbf{38.4 (36.7 / 37.9 / 39.1)}} & 1.092\\
\bottomrule
\end{tabular}
}
\vspace{-.7em}
\caption{\textbf{The deformable module} effectively improves the weakly-supervised learning while introducing negligible computational cost. 'iter time' is the training time per iteration.}
\label{tab:deformable_head} \vspace{-.5em}
\end{table}
%###############################################################################################

%##############################################################################################
% training with more data
\begin{table}[t]
\vspace{-.5em}
\centering
\begin{tabular}{lc}
\toprule
Pretrain-data & AP (r/c/f) \\
\midrule
O365 & {28.6 (24.2 / 27.1 / 30.6)} \\
O365+CC3M & {31.3 (29.4 / 31.7 / 31.3)} \\
O365+GoldG+CC3M & {38.4 (\textbf{36.7} / 37.9 / 39.1)} \\
O365+GoldG+CC15M & {\textbf{40.4} (36.0 / \textbf{41.7} / \textbf{40.0)}} \\
\bottomrule
\end{tabular}
\vspace{-.7em}
\caption{\textbf{Incorporating more data from different sources} consistently improves the performance.}
\label{tab:train_with_more_datasets} \vspace{-.5em}
\end{table}
%####################################################################################

%####################################################################################
% training efficiency
\begin{table}[t]
\vspace{-.5em}
\centering
\resizebox{\linewidth}{!}{
\begin{tabular}{llcc}
\toprule
Model & Pretrain-data & \makecell{Training time \\ (GPU hours)} & \makecell{Training \\ FPS} \\
\midrule
GLIP-T \cite{li2021grounded}\dag & O365+GoldG & 7.4k (3.0k)$^\dag$ & 1.6 \\
DetCLIP-T \cite{yao2022detclip} & O365+GoldG+YFCC1M & \textbf{2.0k} & 4.1 \\
DetCLIPv2-T & O365+GoldG+CC15M &  2.1k & \textbf{25.7} \\
\bottomrule
\end{tabular}
}
\vspace{-.7em}
\caption{\textbf{Training efficiency.} For DetCLIP, we directly use the result reported in the paper; while for GLIP, we calculate the training time based on the FPS provided in the paper. $\dag$: 7.4k is calculated based on the official implementation which trains 30 epochs, while 3.0k is obtained by converting it to our setting of 12 epochs.}
\label{tab:training_efficiency} \vspace{-6mm}
\end{table}
%###############################################################################################

%###########################################################################################
% zero-shot performance
\begin{table*}[t]
\vspace{-4mm}
\centering
\begin{tabular}{lllccc}
\toprule
\multirow{2}{*}{Method} & \multirow{2}{*}{Detector (Backbone)} & \multirow{2}{*}{Pre-Train Data} & \multicolumn{2}{c}{LVIS}  \tabularnewline
 &  &  & {AP} & {AP$_r$ / AP$_c$ / AP$_f$} \\
\midrule
MDETR \cite{kamath2021mdetr} & DETR\cite{carion2020end} (RN101) & GoldG+ & 24.2 & 20.9 / 24.3 / 24.2   \\ 
\midrule
\gc{Supervised} & \gc{ATSS\cite{zhang2020bridging} (Swin-T)} & \gc{LVIS} & \gc{33.6} & \gc{19.7 / 32.4 / 37.2}  \\
GLIP-T\cite{li2021grounded} & DyHead \cite{dai2021dynamic} (Swin-T)  & O365,GoldG,Cap4M & {26.0} & {20.8 / 21.4 / 31.0}  \\
GLIPv2-T \cite{zhang2022glipv2} & DyHead \cite{dai2021dynamic} (Swin-T) & O365,GoldG,Cap4M & {29.0} & { \ \ \ - \ \ / \ \ \  - \ \ \ / \ \ \ -\ \ \ }   \\
DetCLIP-T \cite{yao2022detclip} & ATSS \cite{zhang2020bridging} (Swin-T) & O365,GoldG,YFCC1M & 35.9 & 33.2 / 35.7 / 36.4   \\
\rowcolor{mypink} DetCLIPv2-T (ours) & ATSS \cite{zhang2020bridging} (Swin-T) & O365,GoldG,CC15M & \textbf{40.4} & \textbf{36.0 / 41.7 / 40.0}  \\
\midrule
\gc{Supervised} & \gc{ATSS\cite{zhang2020bridging} (Swin-L)} & \gc{LVIS} & \gc{43.9} & \gc{30.6 / 43.6 / 46.6}  \\
GLIP-L\cite{li2021grounded} & DyHead \cite{dai2021dynamic} (Swin-L) & O365,GoldG,Cap24M & {37.3} & {28.2 / 34.3 / 41.5}  \\
DetCLIP-L \cite{yao2022detclip} & ATSS \cite{zhang2020bridging} (Swin-L) & O365,GoldG,YFCC1M & 38.6 & 36.0 / 38.3 / 39.3  \\
\rowcolor{mypink}DetCLIPv2-L (ours) & ATSS \cite{zhang2020bridging} (Swin-L) & O365,GoldG,CC15M & \textbf{44.7}  &  \textbf{43.1 / 46.3 / 43.7} \\
\bottomrule
\end{tabular}
\vspace{-.7em}
\caption{\textbf{Zero-shot performance} on LVIS minival5k \cite{kamath2021mdetr}. \textit{Fixed} AP \cite{dave2021evaluating} is reported. DetCLIPv2 achieves SoTA performance.}
\label{tab:main_table} \vspace{-1.2em}
\end{table*}
%##########################################################################################

\subsubsection{Ablations for Image-text Contrastive Learning}
\vspace{-1mm}\label{subsubsec:contrastive_ablation}
We investigate key factors for our image-text contrastive learning to work in Table~\ref{tab:contrastive_ablations}. The experiments are conducted with Swin-T-based model on O365+CC3M datasets.

\paragraph{Proposal selection strategy.} 
Selecting representative regions is critical for image-text contrastive learning. Table~\ref{tab:proposal_selection} studies multiple class-agnostic objectness scores for selecting proposals, which includes foreground classfication score \cite{jjfaster2rcnn} (row1), IoU score \cite{jiang2018acquisition} (row2) and centerness \cite{Tian_fcos} (row3). Except for centerness which is originally designed in ATSS \cite{zhang2020bridging}, other 2 scores are predicted by plugging in an additional head after the regression branch. We consider 3 additional scores to utilize textual information: (1) sample-wised text similarity (row4), i.e., each region calculates the similarities with the textual concepts of the sample and the maximum similarity is used as the objectness score; (2) batch-wised text similarity (row5), i.e., the similarities are calculated between a region and textual concepts within a local batch, as described in Sec.~\ref{contrastive_learning}; and (3) multiplying the batch-wised text similarity with the centerness score (row6), which is commonly adopted by conventional detectors \cite{Tian_fcos,zhang2020bridging}. 

Among 3 class-agnostic objectness scores, centerness and IoU scores are superior to classification score, indicating that localization-based objectness scores provide better class-agnostic proposals. The result is consistent with the observations in \cite{kim2022oln}. Considering only the sample-wised text similarity performs worse than using class-agnostic scores, since the regions selected in this way make it easier to distinguish between positive and negative samples in the contrastive learning, thus reducing the learning efficiency. Batch-wise similarity addresses the problem by considering text similarities with negative samples and achieves the best performance of 31.3 AP. Further integrating centerness score results in a performance drop to 30.9 AP.

\paragraph{Word-region alignment strategy.} Table~\ref{tab:matching_strategy} investigates word-region alignment strategies described in Sec.~\ref{contrastive_learning}. Specifically, for fine-grained word-region alignment, 2 matching strategies are studied: (1) 1-to-1 match (row2), i.e.,  each textual concept is matched with its closest region and (2) 1-to-many match (row3), i.e., each textual concept calculates similarities with all regions, which is then aggregated through a softmax-weighted-sum operation. Besides, we also study a coarse-grained image-text matching strategy proposed in \cite{zhou2022detecting} (row1). Specifically, it directly calculates the similarity between the max-size proposal of image and the entire caption of text. Both fine-grained word-region alignment strategies outperform the coarse-grained image-text alignment. Assigning each textual concept with the closest region reaches the best performance (31.3 AP) and substantially saves the GPU memory compared to the 1-to-many strategy, which allows a larger batch size to boost the contrastive learning.

\paragraph{Number of proposals $k$.} Table~\ref{tab:number_proposals} investigates the optimal $k$ when selecting proposals. We vary $k$ from 25 to 200. Using a large $k=200$ results in too many low-quality candidates that slightly decreases the performance. A too small $k=25$ leads to insufficient region extraction which causes a noticeable performance drop. A modest design with 100 proposals achieves the best performance.

\paragraph{Contrastive loss design.} Table~\ref{tab:contrastive_loss} performs ablation experiments on different sides of the image-text contrastive loss (Eq.~\ref{eq:contrastive_loss}). 3 designs are considered: (1). only image-to-text side loss; (2) only text-to-image side loss; and (3) bilateral loss. As discussed in Sec.~\ref{contrastive_learning}, using only image-to-text contrastive loss can lead to a significant performance degradation (29.8 AP) due to the \textit{partial labeling} problem of the image-text pair data. Excluding image-to-text contrastive loss can alleviate the problem and achieving a better performance of 31.3 AP. 

\paragraph{Temperature and Loss weight.} Table~\ref{tab:temperature} and ~\ref{tab:loss_weight} study the optimal values of temperature $\tau$ in Eq.~\ref{eq:contrastive_loss} and loss weight $\lambda$ in Eq.~\ref{eq:overall_objectives}, respectively. The default values of $\lambda=1,\ \tau=0.07$ commonly adopted in standard constastive learning methods  \cite{radford2021learning,yao2021filip} perform poorly in our case. We use $\tau=0.5$ and $\lambda=0.1$ as our final setting. 

% \subsubsection{Comparison with Related Methods}
\vspace{-4mm}
\subsubsection{Effectiveness of Deformable Module} \label{subsubsec:deformable}
\vspace{-1mm}
Table~\ref{tab:deformable_head} studies the effectiveness of the proposed deformable module described in Sec.~\ref{objective_and_model}. The deformable module effectively promotes the  weakly supervised learning. Specifically, it presents negative effect when trained with strongly supervised detection data (row1 and 2), while demonstrating substantial performance improvement when incorporating grounding/image-text pair data without localization supervisions (row3 and 4). Besides, the lightweight deformable module introduces negligible computational cost in terms of training time. 

\vspace{-4mm}
\subsubsection{Incorporating More Data Helps Learning}
\vspace{-1mm}
Table~\ref{tab:train_with_more_datasets} reports the performance gains when scaling up the training data. With the proposed framework, incorporating
more training data from different sources can consistently improve the performance. Compared to training with only Objects365, including CC3M effectively improves the overall AP from 28.6 to 31.3, especially for rare categories (from 24.2 to 29.4, +5.2 AP). GoldG helps significantly improve the overall AP to 38.4 thanks to its instance-level annotations. Including CC12M pushes the envelop further, achieving a 40.4 overall AP which already surpasses the performance of the fully-supervised method (see Table~\ref{tab:main_table}).

\vspace{-3mm}
\subsubsection{Training Efficiency} 
\vspace{-1mm}
We develop DetCLIPv2 with several designs that facilitate training efficiency, including using low-resolution inputs for image-text pairs, limiting the maximum token length of the text encoder to 16, etc. Table~\ref{tab:training_efficiency} compares the training efficiency of DetCLIPv2 with that of GLIP \cite{li2021grounded} and DetCLIP \cite{yao2022detclip}. First, both DetCLIP and DetCLIPv2 are more efficient than GLIP due to the lightweight architecture design, as described in Sec.~\ref{objective_and_model}. 
Besides, DetCLIPv2 is much faster than DetCLIP: it exploits 13$\times$ more image-text pairs than DetCLIP with a similar training time, achieving more than 6$\times$ FPS speed up (25.7 FPS v.s. 4.1 FPS). This indicates the great scaling property of our method and allows a possibility of incorporating a larger-scale image-text pairs to build a more powerful open-vocabulary detection system.

%############################################################################################
\vspace{-1mm}
\subsection{Main Results}\label{subsec:zero-shot_results}
\vspace{-1mm}
\subsubsection{Zero-shot Performance on LVIS}
\vspace{-1mm}
To compare with the existing works, We train DetCLIPv2 with the best setting reported in~\ref{subsubsec:contrastive_ablation}. We vary models' capacity by considering two backbones, i.e., swin-T and swin-L ~\cite{liu2021swin}, denoted as DetCLIPv2-T/L, respectively. Table~\ref{tab:main_table} reports the comparison with MDETR \cite{kamath2021mdetr}, GLIP \cite{li2021grounded}, GLIPv2 \cite{zhang2022glipv2}, and DetCLIP\cite{yao2022detclip} on zero-shot performance. For better demonstration, we also report the performances of the fully-supervised method on LVIS.

DetCLIPv2 outperforms the existing methods by a large margin. Compared to GLIP/GLIPv2, DetCLIPv2 uses a more lightweight backbone (without heavy DyHead \cite{dai2021dynamic} and cross-modal fusion) but still achieves better performances, e.g., DetCLIPv2-T outperforms GLIP-T/GLIPv2-T by 14.4/11.4 AP, respectively. Compared to DetCLIP, DetCLIPv2 achieves 4.5 (40.4 v.s. 35.9) and 6.1 (44.7 v.s. 38.6) AP performance gains for Swin-T- and Swin-L-based models, respectively. Despite using more training data, our total training cost is on par with DetCLIP \cite{yao2022detclip}, as reported in Table~\ref{tab:training_efficiency}.  \textit{Notably, our models beat their fully-supervised conterparts in a zero-shot manner}, e.g, +6.8/0.8 AP for Swin-T- and Swin-L-based models, respectively. Especially, due to the long-tailed property of LVIS, the improvements over rare categories are significant, i.e., more than 10 AP improvements can be observed on both models.

\vspace{-4mm}
\subsubsection{Transfer Results with Fine-tuning}
\vspace{-1mm}
%#########################################################################################
% full-shot performance
\begin{table}[t]
\vspace{-.5em}
\centering
\resizebox{\linewidth}{!}{
\begin{tabular}{lcc}
\toprule
\multirow{2}{*}{Method} & LVIS & ODinW13 \tabularnewline
  & {AP (AP$_{r}$/AP$_{c}$/AP$_{f}$)} & average AP \\
\midrule
GLIP-T \cite{li2021grounded}  & - &  64.9 \\
GLIPv2-T \cite{zhang2022glipv2} & $^\dag$50.6  ( - / - / - )  & 66.5 \\
\rowcolor{mypink} DetCLIPv2-T (ours)  & \textbf{50.7} (44.3/52.4/50.3) & \textbf{68.0} \\
\midrule
GLIP-L \cite{li2021grounded}  & - &  68.9 \\
GLIPv2-B \cite{zhang2022glipv2}  & $^\dag$57.3 ( - / - / - ) &  69.4 \\
GLIPv2-H \cite{zhang2022glipv2}  & $^\dag$59.8 ( - / - / - )  &  \textbf{70.4} \\
\rowcolor{mypink}DetCLIPv2-L (ours)  & \textbf{60.1} (58.3/61.7/59.1) & \textbf{70.4}\\
\bottomrule
\end{tabular}
}
\vspace{-.7em}
\caption{\textbf{Fine-tuning performance}. \textit{Fixed} AP \cite{dave2021evaluating} on LVIS minival5k \cite{kamath2021mdetr} and average AP on ODinW13 \cite{li2021grounded} are reported. Numbers with $^\dag$ mean mask annotation are used for training. }
\label{tab:full-shot_performance} \vspace{-1.3em}
\end{table}
%#########################################################################################

We study the transferability of DetCLIPv2 by fine-tuning it on down-stream tasks. Specifically, we conduct full-shot fine-tuning on LVIS \cite{gupta2019lvis} with 1203 categories and ODinW13 \cite{li2021grounded,zhang2022glipv2} containing 13 detection tasks. The results are shown in Table~\ref{tab:full-shot_performance}. Without using mask annotation for training, DetCLIPv2 slightly outperforms GLIPv2 on LVIS, e.g., 50.7 AP of DetCLIPv2-T v.s. 50.6 AP of GLIPv2-T. On ODinW13, DetCLIPv2-T demonstrates superior performance compared to GLIP-T/GLIPv2-T, outperforming GLIP-T/GLIPv2-T by 3.1/1.5 average AP, respectively; and DetCLIPv2-L with Swin-L backbone achieves the same performance (70.4 average AP) with GLIPV2-H that uses a heavier Swin-H backbone.

\subsection{Visualizations and Analyses}\label{subsec:analysis_and_discussion}
\vspace{-1mm}
\paragraph{Visualization of word-region alignment.} Figure~\ref{fig:word_region_align} visualizes the learning results of word-region alignment on image-text pairs in CC12M \cite{changpinyo2021cc12m}. For each textual concepts, we find its best matching with the highest similarity to it, as described in Sec.~\ref{contrastive_learning}. Our approach achieves accurate word-region alignment (on instance-level) with great generalization, which is demonstrated by several aspects: (1) it successes to recognize concepts that do not covered by detection datasets, e.g., `parsley' in case (b); (2) it works for images with \textit{natural distribution shifts} \cite{taori2020measuring}, e.g., the sketch image in case (a) and the cartoon image in case (e); and (3) it is capable of resolving co-reference expressions, e.g., the `juvenile' in case (h) refers to `young bird' and `curator' in case (c) refers to a person. These capabilities are critical for open-world detectors but cannot be reflected well in the commonly adopted evaluation benchmarks like LVIS \cite{gupta2019lvis}.

%####################################################################################
% average recall
\begin{table}[t]
\vspace{-.5em}
\centering
\begin{tabular}{lc}
\toprule
Pretrain-data & AR (s/m/l) \\
\midrule
O365 & {44.9 (35.2 / 52.9 / 62.0)} \\
O365+GoldG & {57.2 (42.1 / 67.0 / 76.2)} \\
O365+GoldG+CC15M & {\textbf{59.4 (44.5 / 69.4 / 76.8)}} \\
\bottomrule
\end{tabular}
\vspace{-.7em}
\caption{\textbf{Average recall (AR)} across 0.5-0.95 IoU on LVIS. s/m/l denote for small/medium/large objects, respectively.}
\label{tab:average_recall} \vspace{-1.3em}
\end{table}
%####################################################################################

\paragraph{Learning from image-text pairs benefits localization.} Table~\ref{tab:average_recall} provides more evidences showing that learning from image-text pairs also helps localization. Specifically, we evaluate the average recall across 0.5-0.95 IoU on LVIS and compare models trained with different data. Incorporating image-text pairs brings a significant and comprehensive recall improvements for small, medium, and large objects.

\vspace{-1mm}
\section{Conclusion}
\vspace{-1mm}
Learning from massive Internet-crawled data to achieve generic visual/language understanding systems has always
been an important topic for both NLP \cite{brown2020language,devlin2018bert,radford2019language} and CV \cite{radford2021learning,jia2021_align,li2022blip} fields. In this paper, we present DetCLIPv2, a unified end-to-end pre-training framework towards open-vocabulary object detection. By employing a best-matching set similarity between regions and words to guide the contrastive objective, we effectively leverage massive image-text pairs to serve the object detection task. Experiments demonstrate DetCLIPv2's superior open-vocabulary performance and its broad domain coverage. Our method provides a possible way to achieve open-world detection by further scaling up image-text pairs and we leave it to future work.

\renewcommand{\thefootnote}{\arabic{footnote}}
\paragraph{Acknowledgements} We acknowledge the support of MindSpore\footnote{https://www.mindspore.cn/}, CANN (Compute
Architecture for Neural Networks) and Ascend AI Processor used for this research.
\captionsetup[subfigure]{labelformat=simple}

\appendix
\section{Limitations}
Our method provides a possible way to achieve open-world detection by scaling up image-text pairs. However, its localization capability still strongly relies on  bounding box annotations provided in detection data. 
To improve the generalization of localization, designing architectures like \cite{kim2022oln} for robust open-world region proposals is a promising direction for future work. Furthermore, image-text pairs crawled from the Internet are noisy and suffer from severe incomplete descriptions, which undermines the learning efficiency of word-region alignment and requires further designs like \cite{li2022blip} for ameliorating data quality. When further scale up image-text pairs to overwhelm detection data, imbalanced training can potentially hurt the performance, which also calls for a future exploration.

\vspace{-1mm}
\section{More Implementation Details}

In this section, we provide more implementation details for both pre-training and fine-tuning experiments.

\paragraph{Pre-training details.} We pre-train DetCLIPv2 with AdamW \cite{loshchilov2017adamw} optimizer.  The leaning rate first warms up linearly to a peak value (2.8e-4/4e-4 for Swin-B/-L based models, respectively) and then decays following a cosine annealing schedule, where the peak $lr$ values are obtained using a square root scaling rule: $lr=base\_lr\times \sqrt{\frac{\text{batchsize}}{16}}$, where $base\_lr=\text{1e-4}$. We initialize the text encoder with a pre-trained FILIP \cite{yao2021filip} model and reduce the learning rate of text encoder by a factor of 0.1 to preserve the language knowledge obtained in FILIP's pre-training. To save the GPU memory cost and allow a large batch size for contrastive learning, we adopt automatic mixed-precision \cite{micikevicius2018mixed} and gradient checkpointing \cite{ chen2016training} for training. Mmdetection \cite{mmdetection} repository is used for implementation. Table~\ref{tab:pretrain_detail} summarizes the detailed training settings.

%##################################################################################################
% pre-train detail
\begin{table}[t]
\vspace{-.5em}
\centering
\resizebox{\linewidth}{!}{
\begin{tabular}{l|r}
\toprule
Config & Value \\
\midrule
GPUs (V100) & 32(T)/64(L)\\
training epochs & 12 \\
loss weight & $\alpha=1,\beta=2,\lambda=0.1$ \\
optimizer & AdamW \cite{loshchilov2017adamw} \\
optimizer momentum &  $\beta_1=0.9$, $\beta_2=0.999$\\
lr for image encoder & 2.8e-4(T)/4e-4(L) \\
lr for text encoder & 2.8e-5(T)/4e-5(L) \\
weight decay & 0.05 \\
warmup iters & 1000 \\
learning rate schedule & cosine decay \\
batch size (det/grounding) & 128(T)/256(L) \\
batch size (image text pairs) & 6144 \\
input resolution (det/grounding) & $1333\times800$\\
input resolution (image-text pairs) & $320\times320$\\
drop path of visual backbone & 0.2\\
max text token length & 16\\
\# of concepts $M$ (det) & 150 \\
\# of concepts $M$ (grounding) & 100 \\
label smooth for contrastive loss & 0.1\\
augmentation & \makecell[r]{ multi-scale training,\\ random flip}\\
\bottomrule
\end{tabular}
}
\vspace{-.5em}
\caption{\textbf{Detailed pre-training settings} of DetCLIPv2. T/L in parentheses denote Swin-T/L models, respectively. Det/grounding mean detection and grounding data, respectively.}
\label{tab:pretrain_detail}
\vspace{-4mm}
\end{table}

%##################################################################################################
\paragraph{Fine-tuning details.} We fine-tune DetCLIPv2 on 2 datasets, i.e., LVIS \cite{gupta2019lvis} and ODinW13 \cite{li2021grounded}. For LVIS, we follow most settings of pre-training except that we use a smaller learning rate and the total epochs are set to 24 (i.e., 2x schedule).  Table~\ref{tab:finetune_lvis_details} summarizes the detailed setting of fine-tuning LVIS. For ODinW13, since the number of training samples of different datasets varies a lot, we cannot set the same training epoch for all datasets. To avoid tedious hyper-parameter turning and ensure a sufficient training for all datasets, we adopt a long training schedule with early stop mechanism. Specifically, we assign a maximum training epoch with an auto-step learning rate schedule. We monitor the performance and decay the learning rate by 0.1 when the performance reaches a plateau for a tolerance of $t_1$ epochs. If the learning rate reaches a given minimum value and there is no performance improvement for $t_2$ epochs, the training exits. We use the same learning rate configuration for all datasets and \textit{do not} search optimal hyper-parameters for each dataset separately.

%##################################################################################################
% lvis finetune detail
\begin{table}[t]
\centering
\resizebox{\linewidth}{!}{
\begin{tabular}{l|r}
\toprule
Config & Value \\
\midrule
GPUs (V100) & 16 \\
training epochs & 24 \\
optimizer & AdamW \cite{loshchilov2017adamw} \\
optimizer momentum &  $\beta_1=0.9$, $\beta_2=0.999$ \\
lr for image encoder & 4e-5 \\
lr for text encoder & 4e-6 \\
weight decay & 0.05 \\
warmup iters & 1000 \\
learning rate schedule & cosine decay \\
batch size  & 64 \\
input resolution  & $1333\times800$\\
drop path of visual backbone & 0.2\\
\# of concepts $M$ & 150\\
augmentation & \makecell[r]{ multi-scale training,\\ random flip} \\
\bottomrule
\end{tabular}
}
\vspace{-.5em}
\caption{\textbf{Detailed fine-tuning settings} for LVIS \cite{gupta2019lvis}.}
\label{tab:finetune_lvis_details}
\vspace{-.5em}
\end{table}
%##################################################################################################

%##################################################################################################
% odinw13 finetune detail
\begin{table}[t]
\centering
\resizebox{\linewidth}{!}{
\begin{tabular}{l|r}
\toprule
Config & Value \\
\midrule
GPUs (V100) & 8\\
maximum training epochs & 250 \\
optimizer & AdamW \cite{loshchilov2017adamw} \\
optimizer momentum &  $\beta_1=0.9$, $\beta_2=0.999$\\
lr for image encoder & 4e-5 \\
lr for text encoder & 4e-7 \\
weight decay & 0.05 \\
warmup iters & 500 \\
learning rate schedule & auto-step decay \\
lr decay tolerance $t_1$ (epochs) & 5 \\
training exit tolerance $t_2$ (epochs) & 8 \\
minimum lr to stop decay & 1e-8 \\
batch size  & 32 \\
input resolution  & $1333\times800$\\
drop path of visual backbone & 0.2\\
augmentation & \makecell[r]{ multi-scale training,\\ random flip}\\
\bottomrule
\end{tabular}
}
\vspace{-.5em}
\caption{\textbf{Detailed fine-tuning settings} for ODinW13 \cite{li2021grounded}.}
\label{tab:finetune_lvis_details}
\vspace{-.5em}
\end{table}
%##################################################################################################
\vspace{-1mm}
\section{More Experimental Results} \label{sec:experiments_appendix}
\vspace{-1mm}
\paragraph{Effect of input resolution of image-text pairs.} 
Reducing the input resolution of massive image-text pairs can significantly boost the  training efficiency while may lead to performance degradation. Table~\ref{tab:input_resolution} studies the effect of input resolution change of image-text pairs, where we conduct experiments with the Swin-T-based model on O365+CC3M and vary the resolution of CC3M data from $224\times224$ to $384\times384$. Increasing the resolution from $256$ to $320$ leads to an obvious performance improvement (from 30.5 AP to 31.5 AP). However, further increasing it to $384$ only brings limited performance gains and introduces considerable memory and training time overhead. Therefore, we choose $320\times320$ as our final setting.

%####################################################################################
% resolution of image-text pairs
\begin{table}[t]
\centering
\begin{tabular}{lccc}
\toprule
Input res & GPU Memory & \makecell{Training time\\(GPU hours)}  & AP \\
\midrule
$224\times224$ & 14.1 GB & 697.6 & 30.5  \\
$256\times256$ & 16.0 GB & 729.6 & 30.5 \\
\rowcolor{mygray}$320\times320$ & 20.7 GB & 793.6 & 31.3 \\
$384\times384$ & 24.2 GB & 876.8 & 31.5 \\
\bottomrule
\end{tabular}
\caption{\textbf{Input resolution change of image-text pairs.} We use Swin-T-based model trained with O365+CC3M. Zero-shot AP on LVIS minival5k is reported. Using resolution of $320\times320$ (marked in \colorbox{mygray}{gray}) achieves the best trade-off between computational cost and model performance.}
\label{tab:input_resolution} 
\vspace{-3mm}
\end{table}
%####################################################################################

%####################################################################################
% + imagenet 
\begin{table}[t]
\centering
\resizebox{\linewidth}{!}{
\begin{tabular}{lllc}
\toprule
\# & Backbone & Pretrain-data & AP (r/c/f) \\
\midrule
1 & Swin-T& O365 &  28.6 (24.2/27.1/30.6) \\
2 & Swin-T& O365+IN1k &  30.4 (\textbf{32.2}/29.4/30.9) \\
3 & Swin-T& O365+CC3M & \textbf{31.3} (29.4/\textbf{31.7}/\textbf{31.3}) \\
\midrule
4 & Swin-T& O365+GoldG+CC15M & 40.4 (36.0/41.7/\textbf{40.0}) \\
5 & Swin-T& \makecell[l]{O365+GoldG+CC15M\\+IN1k}  & \textbf{40.6} (\textbf{38.2}/\textbf{42.0}/39.9) \\
\midrule
6 & Swin-L& O365+GoldG+CC15M & \textbf{44.7} (43.1/\textbf{46.3}/43.7) \\
7 & Swin-L& \makecell[l]{O365+GoldG+CC15M\\+IN1k} & \textbf{44.7} (\textbf{43.8}/45.4/\textbf{44.3}) \\
\bottomrule
\end{tabular}
}
\caption{\textbf{Effect of incorporating IN21k data for training}. r/c/f indicate rare/common/frequent categories, respectively. Zero-shot AP on LVIS minival5k is reported. }
\vspace{-2mm}
\label{tab:train_with_imagenet} 
\end{table}
%####################################################################################

%###########################################################################################
% zero-shot performance
\begin{table*}[th]
\vspace{-1mm}
\centering
\begin{tabular}{llcccc}
\toprule
\multirow{2}{*}{Method} & \multirow{2}{*}{Detector (Backbone)} & \multicolumn{2}{c}{LVIS (zero-shot)} & \multicolumn{2}{c}{LVIS (fine-tune)}  \\
 &  & {AP} & {AP$_r$ / AP$_c$ / AP$_f$} & {AP} & {AP$_r$ / AP$_c$ / AP$_f$} \\
\midrule
\gc{Supervised} & \gc{ATSS\cite{zhang2020bridging} (Swin-T)} & \gc{ - } & \gc{ - / - / - } & \gc{28.4} & \gc{18.9 / 27.3 / 33.6}  \\
GLIP-T\cite{li2021grounded} & DyHead \cite{dai2021dynamic} (Swin-T) & 17.2 & {10.1 / 12.5 / 25.2} & {-} & { - / - / - }  \\
DetCLIP-T \cite{yao2022detclip} & ATSS \cite{zhang2020bridging} (Swin-T) & 28.4 & {25.0 / 27.0 / 31.6 } & {-} & { - / - / - }   \\
\rowcolor{mypink} DetCLIPv2-T (ours) & ATSS \cite{zhang2020bridging} (Swin-T) & \textbf{32.8} & \textbf{31.0 / 31.7 / 34.8} & \textbf{43.7} & \textbf{40.2 / 42.7 / 46.3}  \\
\midrule
\gc{Supervised} & \gc{ATSS\cite{zhang2020bridging} (Swin-L)} & \gc{ - } & \gc{ - / - / - } & \gc{38.3} & \gc{28.5 / 38.1 / 42.9}  \\
GLIP-L\cite{li2021grounded} & DyHead \cite{dai2021dynamic} (Swin-L) & 26.9 & {17.1 / 23.3 / 36.4} & {-} & { - / - / - }  \\
DetCLIP-L \cite{yao2022detclip} & ATSS \cite{zhang2020bridging} (Swin-L) & 31.2 & { 27.6 / 29.6 / 34.5 } & {-} & { - / - / - }  \\
\rowcolor{mypink}DetCLIPv2-L (ours) & ATSS \cite{zhang2020bridging} (Swin-L) & \textbf{36.6} & \textbf{33.3 / 36.2 / 38.5} & \textbf{53.1} &  \textbf{49.0 / 53.2 / 54.9 } \\
\bottomrule
\end{tabular}
\vspace{-.5em}
\caption{\textbf{Performance on LVIS \cite{kamath2021mdetr} val split}. \textit{Fixed} AP \cite{dave2021evaluating} is reported. DetCLIPv2 achieves SoTA performance.}
\vspace{-.5em}
\label{tab:lvis_val}
\end{table*}
%##########################################################################################

%##########################################################################################
% detailed odinw13 finetune results
\begin{table*}[th]
\centering
\begin{tabular}{lccccccc}
\toprule 
{Model} & PascalVOC & AerialDrone & Aquarium & Rabbits & EgoHands & Mushrooms & Packages  \\
\midrule
{GLIP-T} & 62.3 & 31.2 & 52.5 & 70.8 & 78.7 & 88.1 & 75.6 \\
{GLIPv2-T} & 66.4 & 30.2 & 52.5 & 74.8 & 80.0 & 88.1 & 74.3 \\
\rowcolor{mypink}{DetCLIPv2-T (ours)} & 67.5 & 41.8 & 50.8 & 80.4 & 79.8 & 90.1 & 73.7 \\
\midrule
{GLIP-L} & 69.6 & 32.6 & 56.6 & 76.4 & 79.4 & 88.1 & 67.1 \\
{GLIPv2-B} & 71.1 & 32.6 & 57.5 & 73.6 & 80.0 & 88.1 & 74.9 \\
{GLIPv2-H} & 74.4 & 36.3 & 58.7 & 77.1 & 79.3 & 88.1 & 74.3 \\
\rowcolor{mypink}{DetCLIPv2-L (ours)} & 74.4 & 44.1 & 54.7 & 80.9 & 79.9 & 90 & 74.1 \\
\bottomrule
\toprule
{Model} &  Raccoon & Shellfish & Vehicles & Pistols & Pothole & Thermal & Avg \\
\midrule
{GLIP-T} & 61.4 & 51.4 & 65.3 & 71.2 & 58.7 & 76.7 & 64.9 \\
{GLIPv2-T} & 63.7 & 54.4 & 63.0 & 73.0 & 60.1 & 83.5 & 66.5 \\
\rowcolor{mypink}{DetCLIPv2-T (ours)} & 70.8 & 54.8 & 66.5 & 77.7 & 54.8 & 82.2 & \textbf{68.5}  \\
\midrule
{GLIP-L} & 69.4 & 65.8 & 71.6 & 75.7 & 60.3 & 83.1 & 68.9 \\
{GLIPv2-B} & 68.2 & 70.6 & 71.2 & 76.5 & 58.7 & 79.6 & 69.4 \\
{GLIPv2-H} & 73.1 & 70.0 & 72.2 & 72.5 & 58.3 & 81.4 & \textbf{70.4} \\
\rowcolor{mypink}{DetCLIPv2-L (ours)} & 69.4 & 61.2 & 68.1 & 80.3 & 57.1 & 81.1 & \textbf{70.4} \\
\bottomrule
\end{tabular}
\vspace{-.5em}
\caption{\textbf{Detailed fine-tuning AP (\%) performance on ODinW13.}}
\label{table:odinw13_finetune_results}
\vspace{-.8em}
\end{table*}
%##########################################################################################

\paragraph{Incorporating classification dataset.} 
 Our framework can be viewed as a more general design for weakly-supervised (WSOD) approaches, which eliminates the limit of pre-defined categories in traditional WSOD methods. By formulating classification data as a special type of image-text pair data, our method is capable of incorporating it into training. 
Specifically,  we use the category name as the caption for each image. To select region proposals, similar to image-text pair, we collect category names in a batch to calculate the similarity with a region and select the maximum value as the objectness score. Considering classification image typically contains only 1 main object, we select top $k=1$ proposal. Finally, the contrastive loss for image-text pair is replaced with the cross entropy loss for classification, and we also set loss weight $\lambda=0.1$.

We perform experiments on ImageNet1k \cite{deng2009imagenet} (denoted as IN1k) and show the results in Table~\ref{tab:train_with_imagenet}.  2 settings are considered: 1. we train IN1k with only the detection data, i.e., O365; and 2. we incorporate IN1k into the final version of DetCLIPv2, i.e., all data including O365, CC15M, GoldG and IN1k are used. During training, we use the same image-text pair setting for IN1k and replicate IN1k by 3 times to make it have a similar size to CC3M.

First, incorporating classification data when using only detection data can significantly improve the performance from 28.6 to 30.4 (rows 1 and 2), especially for rare categories (from 24.4 to 32.2 AP, +8 AP), yet it is slightly worse than using CC3M in terms of overall AP (rows 2 and 3). However, when all data are used, the advantage of IN1k diminishes, i.e., it brings only 0.2 overall AP for Swin-T-based model (rows 4 and 5) and no performance gain is observed for  Swin-L-based model (rows 6 and 7). Therefore, we exclude the classification data from our method to keep it as neat as possible. 

\paragraph{More results on LVIS.} To make a comprehensive comparison with the existing methods, we also evaluate DetCLIPv2 with the complete validation set of LVIS \cite{gupta2019lvis} (including 20k images with 1203 categories), on both zero-shot and fine-tuning settings. Table~\ref{tab:lvis_val} exhibits the results. DetCLIPv2 outperforms GLIP and DetCLIP by a large margin for both T/L models, e.g., DetCLIP-T surpasses GLIP-T/DetCLIP-T by 15.6/4.4 AP, respectively. Besides, by pre-training on large-scale hybrid data and  fine-tuning on LVIS,  DetCLIPv2 achieves significant improvements over the fully-supervised method, i.e., about 15 overall AP improvement can be observed for both T/L models.

\paragraph{Detailed fine-tuning results for ODinW13.}
Table~\ref{table:odinw13_finetune_results} reports the detailed fine-tuning performance for 13 datasets contained in ODinW13, and we make a comparison between DetCLIPv2 and GLIP \cite{li2021grounded}/GLIPv2 \cite{zhang2022glipv2}. DetCLIPv2-L/T surpass their GLIP\cite{li2021grounded}/GLIPv2\cite{zhang2022glipv2} counterparts on average AP over 13 datasets.

\vspace{-1mm}
\section{More Visualization Results}
Figure~\ref{fig:word-region-vis-1} and~\ref{fig:word-region-vis-2} provide more visualization examples of word-region alignment learnt by DetCLIPv2, using the images from CC12M. As mentioned in the main paper, we find the optimal-match region in the image for each textual concept in the caption. As can be seen, DetCLIPv2 learns to recognize and locate various concepts with broad domain coverage, including comic objects (i.e., Monkey King,  Santa Claus, etc.), abstract concepts (i.e., `man's best friend' means a dog) and many concepts that are not covered by the detection/grounding data (i.e., tensioner, tattoos, lifebuoy and etc.), which demonstrates the effectiveness of learning from massive image-text pairs.

%#################################################################
% word-region alignment figures
\renewcommand{\thefigure}{1-a}
\begin{figure*}[ht!]
\vspace{-4mm}
\begin{center}
\includegraphics[width=\textwidth]{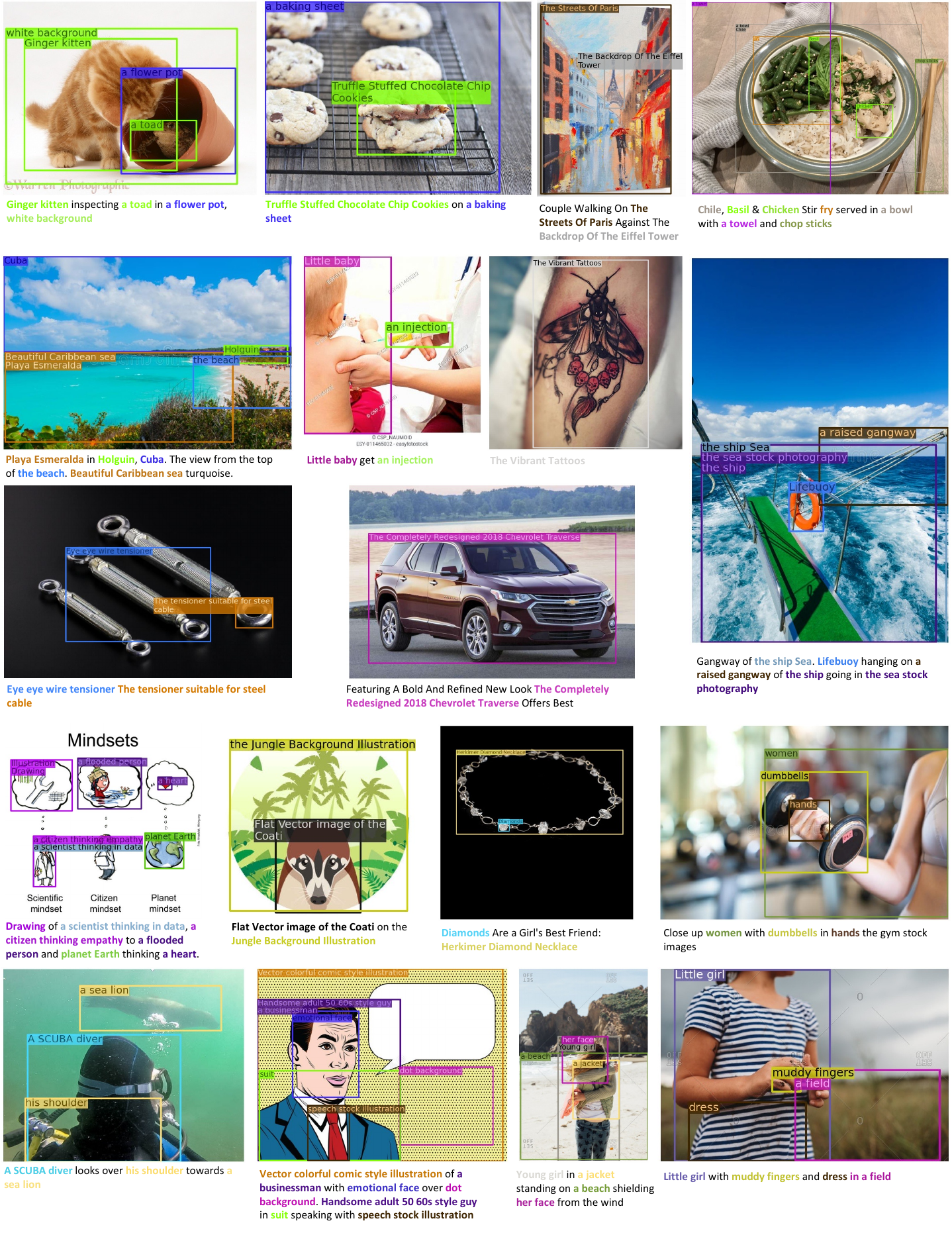}
\vspace{-6mm}
\caption{\textbf{More visualizations for word-region alignment}. DetCLIPv2 learns word-region alignment with broad domain coverage.}
\label{fig:word-region-vis-1}
\end{center}
\vspace{-7mm}
\end{figure*}

\renewcommand{\thefigure}{1-b}
\begin{figure*}[ht!]
\vspace{-4mm}
\centering
\includegraphics[width=\textwidth]{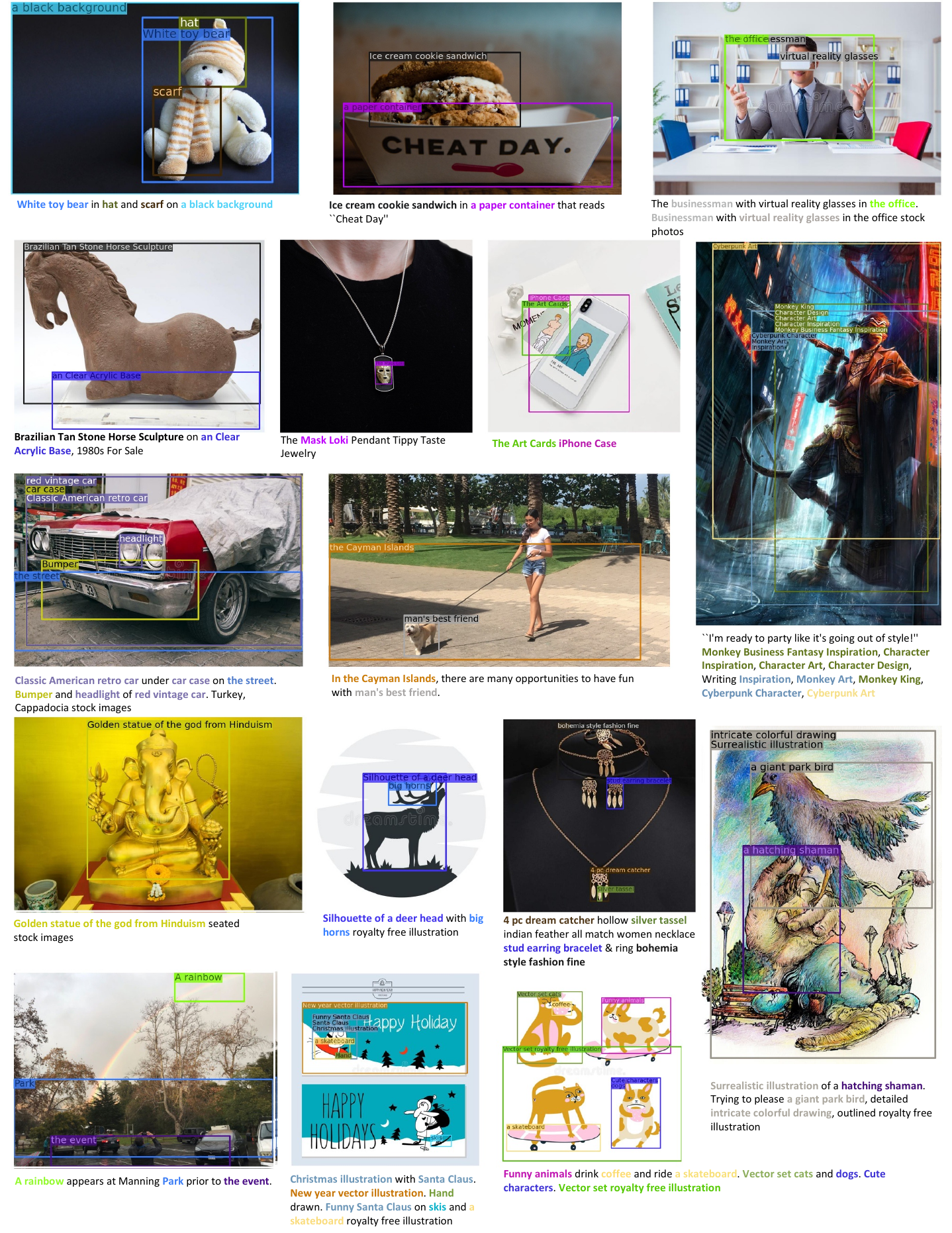}
\vspace{-7mm}
\caption{\textbf{More visualizations for word-region alignment (cont.)}. DetCLIPv2 learns word-region alignment with broad domain coverage.}
\label{fig:word-region-vis-2}
\vspace{-7mm}
\end{figure*}

\setcounter{figure}{2}
\renewcommand{\thefigure}{\arabic{figure}}
%#################################################################

%%%%%%%%% REFERENCES
\newpage
\newpage
{\small
\bibliographystyle{ieee_fullname}
\bibliography{egbib}
}

\end{document}